# What we really want to find by Sentiment Analysis: The Relationship between Computational Models and Psychological State

**Hwiyeol Jo (hwiyeolj@gmail.com)**
Department of Computer Science & Engineering, Seoul National University,
Seoul, 151-742, Republic of Korea

**Soo-Min Kim (spdlqj438@naver.com)**
Department of Psychology, Chung-Ang University
Seoul, 110-745, Republic of Korea

**Jeong Ryu (ryujeong@gmail.com)**
Department of Psychology, Yonsei University
Seoul, 120-749, Republic of Korea

**Abstract**

As the first step to model emotional state of a person, we build sentiment analysis models with existing deep neural network algorithms and compare the models with psychological measurements to enlighten the relationship. In the experiments, we first examined psychological state of 64 participants and asked them to summarize the story of a book, Chronicle of a Death Foretold (Márquez, 1981). Secondly, we trained models using crawled 365,802 movie review data; then we evaluated participants' summaries using the pre-trained model as a concept of transfer learning. With the background that emotion affects on memories, we investigated the relationship between the evaluation score of the summaries from computational models and the examined psychological measurements. The result shows that although CNN performed the best among other deep neural network algorithms (LSTM, GRU), its results are not related to the psychological state. Rather, GRU shows more explainable results depending on the psychological state. The contribution of this paper can be summarized as follows: (1) we enlighten the relationship between computational models and psychological measurements. (2) we suggest this framework as objective methods to evaluate the emotion; the real sentiment analysis of a person.

## Introduction

Sentiment analysis is a kind of text mining, which is to predict human mind, specifically the emotional state of a person by extracting specific emotional expressions from the text (Pang & Lee, 2008). Analyzing product reviews can be an example. The reviews include positive or negative words about the product. With the process of automatic classification or scoring of those data, we can apply the result to determine whether the product is attractive or not for customers. In the field of Natural Language Processing (NLP), the sentiment analysis problem is hard to achieve both high accuracy and performance. One of the most commonly used datasets in sentiment analysis is Stanford Sentient Treebank (SST), which consists of movie 5 classes labeled reviews (very negative, negative, neutral, positive, very positive) (Socher et al., 2013). One of recent model handling the dataset is Dynamic Memory Network (Kumar et al., 2015). Although the model performs well in binary classification (positive or negative), its fine-grained accuracy is only 51.2%, which is considerably low.

As we mentioned, recent sentiment analysis models have tried to determine the polarity and intensiveness of data. The algorithm divides the text data by polarity and learns the meaning of words which includes each polarity. However, this work is started with a question, "What we really want to find by sentiment analysis?" Although previous studies have solved various sentiment analysis problems by designing its architectures and algorithms, the model should be able to predict emotional state of a person, not the text data. Studies should focus on finding the sentiment of the individual who wrote the words, not on the emotional arousal of words or sentences. The different viewpoint to sentiment analysis is along with strong Artificial Intelligence, which research goal is to imitate human intelligence and the human mind.

As the first step to model emotional state of a person, we build sentiment analysis models with existing deep neural network algorithms and compare the models with psychological measurements to enlighten the relationship between them. With this framework, we could suggest a new framework to evaluate emotion or psychological state. Research which studied the emotion try to assess the emotion using only measurements (e.g. PANAS), otherwise determined by subjective methods. For example, Mayer & Geher (1996) employed group consensus to identify the emotion, Diener et al. (1985) computed the intensity of emotion by the difference between positive scale in a day and mean positive scale. The studies were unsuccessful in representing a method to determine what the emotion and how its intensity in objective way without psychological measurements.

To show the difference in retrieved a memory under the equal circumstance, we have used the 'Chronicle of a Death Foretold.' The novel has been used in diverse fields of study, ranging from literature and language (Aghaei & Hayati, 2014), cognitive science (Popova, 2015), psychology (Scheper-Hughes, 2003), culture (Weiss, 1990) and so on.

Finding which emotion (or psychological state) the sentiment analysis model with the existing deep learning algorithm can predict, we hope that our results can make sentiment analysis researchers see the final goal of their works, which is to identify the emotional state of the person, not the words. If this approach is thought to be reasonable, this research might be a work to suggest a new framework to evaluate sentiment analysis model.

Now, we will briefly summarize the background knowledge needed for understanding our method, which consists of our experimental data and training data of sentiment analysis model each other. Afterward, we will introduce our overall experimental method, explain the results, and draw a conclusion with a discussion.

## Backgrounds

### Emotion effects on Memory

Emotion affects the contents of memories both when we store and retrieve the memory. Theory "Affect as information framework" explains that if we are positive, we tend to be interpretive, relational processing while being detailed, stimulus-bound, referential processing in negative affective (Clore et al., 2001). Kensinger et al. (2007a) found the different specificity of an event. Emotion can consolidate memory. Emotional memories are kept fairly untouched and slowly forgotten because they were repeatedly rehearsed. The negative emotion can improve an accuracy of the memory (Kensinger, 2007b). Research showed that emotion could make affect regulation (Raes et al., 2003; Fonagy et al., 2004). If an event has occurred for a long period, people evaluate the event depending on their emotion when the emotional arousal is peak and is at the end of the event. Mood congruent information which is in the same valence with one's present mood would be more perceived, memorized, retrieved, and utilized to decide than mood incongruent information.

### Chronicle of a Death Foretold

The novel was written by Gabriel Garcia Márquez (1981), Chronicle of a death foretold, came from his experience of seeing a murder case. In a small village near the coast, Vicario brothers decided to kill Nasar with unconvincing evidences that Nasar raped their sister Angela, to regain their family's honor by honor killing. Even though the brothers notified the time, the place, and the reason they will kill him, nobody warned Narsar against the threat. The village became full of honor with revenge, violence in indifference, and false witness in misunderstandings, as the story drove into tragedy.

For the distinctive characteristics of the novel (e.g. honor killing, the indifference of the crowd, narrative style, and non-sequential storytelling), it has been studied for a long time.

### Deep Neural Networks

Neural Networks was inspired by the human brain activity, imitating logical calculus of nervous activity. Traditional Neural Networks showed powerful performance in accuracy. However, it had lots of limitations such as complexity, overfitting problem. Through various studies in neural network and the improvement of hardware, the deep neural networks, which is also called deep learning, start with stacking more and more hidden nodes and hidden layers. The core idea of the deep neural networks are to find important features, and patterns then use it to prediction or classification. Recent popular deep neural networks consist of three main algorithms: CNN; Convolutional Neural Network (LeCun & Bengio, 1995), RNN; Recurrent Neural Networks as Long-Short Term Memory (Hochreiter & Schmidhuber, 1997) and Gated Recurrent Units (Cho et al., 2014).

### Transfer Learning

Transfer learning is a technique to convey knowledge from other tasks, which are related to the target task (Pan & Yang, 2010) likewise human learning. It has an advantage in that requires fewer amounts of training data because of employing the knowledge constructed from similar problems.

Transfer learning takes three steps: First, the framework learns from the source. Secondly, the framework transfers the knowledge from source to target, and the last is to learn from the target. While traditional machine learning concentrates on solving a problem using training data in a domain, the data from various domains can be used in transfer learning.

## Data

### Experimental Data

**Participants Information**

All the participants were university students in the Republic of Korea. Evenly distributed participants were sampled; 39 (60.94%) male and 25 (39.06%) female and 31 (48.44%) students majored Liberal Arts while the others (51.56%) majored in Science. The average and standard deviation of age is 22.5091 and 2.4257, respectively.

The psychological state; affectivity (Positive Affective Negative Affective Schedule; PANAS) and depression (Center for Epidemiological Studies – Depression; CES-D), of the undergraduate students is presented in Table 1.

Table 1: Psychological state of participants

| Characteristic | N* | % | Mean (SD) | Range |
|---|---|---|---|---|
| PANAS** | 64 | | | |
| Positive | | | 33.594 (6.592) | 22-50 |
| Negative | | | 28.828 (7.629) | 15-46 |
| CES-D*** | 63**** | | 26.635 (12.207) | 5-50 |

| | | |
|---|---|---|
| Depression (>=21) | 41 | 65.08 |
| Non-depression (<21) | 22 | 34.92 |

\* The number of subjects
\*\* Positive Affective Negative Affective Schedule
\*\*\* Center for Epidemiological Studies – Depression
\*\*\*\* A missing data; a participant did not check CES-D

### Train Data

We collected movie review data from Naver, the most popular Korean portal site. We have implemented web crawling tool using Scrappy, a Python package. The review data includes the information of titles, scores, and comments on the movie. The distribution of scores in the review data is presented in Table 2.

However, we can find the distribution of scores is biased on 10 points. Some of the reason can be a fake review for advertising or getting the point by leaving formal reviews. Therefore, we decided to ignore 10 point review data with the expectation that 8 or 9 point reviews might include positive expressions as much as 10 point reviews.

Table 2: The distribution of scores of the review data

| Score | N* | % | Score | N | % |
|---|---|---|---|---|---|
| 1 | 61,307 | 16.76 | 6 | 26,220 | 7.17 |
| 2 | 8,700 | 2.38 | 7 | 44,736 | 12.22 |
| 3 | 8,674 | 2.37 | 8 | 89,310 | 24.41 |
| 4 | 9,223 | 2.52 | 9 | 97,169 | 26.56 |
| 5 | 20,463 | 5.59 | ~~10~~ | ~~327,544~~ | |
| | | [Total – 365,802] | | | |

\* The number of data

### Method

The subjects of the experiment were undergraduate students (N=64) who enrolled in "Introduction to Cognitive Science" class in Konkuk University, Republic of Korea, during 2014 fall semester. 64 students are applied in the beginning, the end, only 55 complete all the experiments, measuring psychological status and summarizing the story of the book.

We examined depression (Center for Epidemiological Studies – Depression, Radloff, 1977), present positive affective and negative affective (Positive Affective and Negative Affective Schedule, Watson et al., 1988). CES-D score shows how depression is severe during specific periods. The cut-off point for depression, 16 is suggested by Radloff, whereas Cho et al. (1993) suggested 21 for epidemiological studies for Koreans. PANAS measures positive affectivity (interested, excited, strong, enthusiastic, proud, alert, inspired, determined, attentive, and active) and negative affectivity (distressed, upset, guilty, scared, hostile, irritable, ashamed, nervous, jittery, and afraid).

Participants first completed their psychological state; CES-D, PANAS as well as personal factors; Life Orientation Test (Scheier & Carver, 1985), Core Self Evaluation Scale (Judge et al., 2003), Social Support (Sarason et al., 1983). However, the personal factors could not make a meaningful result. As for their assignments, we gave them a book, 'Chronicle of a Death Foretold' (Márquez, 1981) to read carefully, announcing that they will take a quiz in the following week. In the next week, the students were asked to summarize the content of the book as specific as they could remember.

We first scrapped movie review data (as stated above) and implemented a simple sentiment analysis model based on existing deep neural networks, CNN (Convolutional Neural Networks, LeCun & Bengio, 1995), LSTM (Long-Short Term Memory, Hochreiter & Schmidhuber, 1997), and GRU (Gated Recurrent Units, Cho et al., 2014). We followed the general approach (Pang & Lee, 2008) that regards the sentiment analysis as a classification problem which class is 1 to 10 (in this case, 1 to 9). We trained the model using the movie data. Then, we input the summaries of the book to each model to evaluate the emotional expressions included in their summaries.

We have obtained the concept of transfer learning, which is that the knowledge from similar tasks can be used for doing target tasks. We hypothesize that movie reviews have emotional words as well as the intense of the word from 1 to 10 (in this case, 1 to 9) so they can be sources to evaluate emotional contents of the summaries.

## Results

### Model Accuracy

For an experiment, we collected 365,802 movie review data, allocating 70%, 15%, 15% of the data for train, validation, and test, respectively.

### 9-Level polarity

We divided the score of movie reviews from 1 to 9. The accuracy is presented in Table 3.

Table 3: The accuracy of our model as 9-level polarity

| Model | Test Accuracy (Top-1,3,5) | | |
|---|---|---|---|
| CNN* | 0.4132 | 0.7851 | 0.9049 |
| LSTM** | 0.4035 | 0.7700 | 0.8923 |
| GRU*** | 0.4092 | 0.7767 | 0.9001 |

\* Convolutional Neural Networks
\*\* Long-Short Term Memory
\*\*\* Gated Recurrent Units

### 3-Level polarity

We divided the score of movie reviews by 3-level polarity, which is negative, neutral, or positive. We merged 1, 2, and 3 into negative, 4, 5, and the other (7, 8, 9) into positive. The accuracy is presented in Table 4.

Table 4: The accuracy of our model as 3-level polarity

| Model | Test Accuracy |
|---|---|
| CNN* | 0.7460 |
| LSTM** | 0.7360 |

| | GRU*** | | | 0.7402 | |

\* Convolutional Neural Networks
\*\* Long-Short Term Memory
\*\*\* Gated Recurrent Units

## Evaluation Results of the Summaries

The three models generated different evaluation results. When evaluating the summaries with CNN model, 3 summaries were 1 point, 4 of them were 2 point, 2 of them were 3 point, 3 of them were 4 point, 1 of them was 5 point, 29 of them were 6 point, 11 of them were 7 point, and 2 of them were 8 point. With LSTM, 24 of them were 6 points, and 31 of them were 7 points. With GRU, 41 of them were 3 points, 8 of them were 6 points, and 6 of them were 7 points.

## Relationship between Computational Models and Psychological State

With the belief that the scoring is reasonable, we investigated the relationship between the scores in 9-level polarity and psychological state. This analysis method is based on studies that the content of retrieved memory depends on the emotional state (Bower, 1981; Blaney, 1986).

First, we divided the participants by PANAS. Using cut-off point as the average value, we performed *t*-test to investigate the relationship between evaluation scores with PANAS. We defined *PANAS_SUM* as the sum of positive PANAS and negative PANAS to indicate how the person is emotionally expressive. The *t*-test results are presented in Table 5.

Table 5: *t*-test results when dividing the participants by PANAS average value

| Measurement | Model | Characteristic | N | Mean (SD) | *t* or *F*-value |
|---|---|---|---|---|---|
| PANAS Positive | CNN | More Positive (>=34.1) | 26 | 5.692 (1.850) | $T=.858$ $p=.395$ |
| | | Less Positive (<34.1) | 29 | 5.276 (1.750) | |
| | LSTM | More Positive (>=34.1) | 26 | 6.538 (0.508) | $T=-.350$ $p=.727$ |
| | | Less Positive (<34.1) | 29 | 6.586 (0.501) | |
| | GRU | More Positive (>=34.1) | 26 | 3.538 (1.303) | $F=9.286$ $p=.121$ |
| | | Less Positive (<34.1) | 29 | 4.172 (1.671) | |
| PANAS Negative | CNN | More Negative (>=27.8) | 28 | 5.571 (1.752) | $T=.412$ $p=.682$ |
| | | Less Negative (<27.8) | 27 | 5.370 (1.864) | |
| | LSTM | More Negative (>=27.8) | 28 | 6.536 (0.508) | $T=-.418$ $p=.678$ |
| | | Less Negative (<27.8) | 27 | 6.593 (0.501) | |
| | GRU | More Negative (>=27.8) | 28 | 4.071 (1.609) | $T=.982$ $p=.331$ |
| | | Less Negative (<27.8) | 27 | 3.667 (1.441) | |
| PANAS SUM | CNN | MoreExpressive (>=61.9) | 26 | 5.500 (1.881) | $T=.106$ $p=.916$ |
| | | LessExpressive (<61.9) | 29 | 5.448 (1.744) | |
| | LSTM | MoreExpressive (>=61.9) | 26 | 6.654 (0.485) | $T=1.273$ $p=.209$ |
| | | LessExpressive (<61.9) | 29 | 6.483 (0.509) | |
| | GRU | MoreExpressive (>=61.9) | 26 | 3.231 (0.815) | $F=67.33$ ***p=.002*** |
| | | LessExpressive (<61.9) | 29 | 4.448 (1.785) | |

We could observe that GRU can distinguish a difference between the expressive person (M=3.231, SD=0.815) and less expressive person (M=4.448, SD=1.785); $F(67.33)=-3.309, p=.002$. The result is reasonable that near 4.5 point means less positive or negative than near 1 point or 9 points.

Secondly, we divided the participants by CES-D. Using cut-off point as the average value, we performed *t*-test as the same. The *t*-test result is presented in Table 6.

Table 6: *t*-test results when dividing the participants by CES-D average value

| Measurement | Model | Characteristic | N | Mean (SD) | *t* or *F*-value |
|---|---|---|---|---|---|
| CES-D | CNN | More Depressed (>=26.1) | 24 | 6.000 (1.216) | $F=13.32$ ***p=.041*** |
| | | Less Depressed (<26.1) | 31 | 5.065 (2.065) | |
| | LSTM | More Depressed (>=26.1) | 24 | 6.542 (0.509) | $T=-.284$ $p=.778$ |
| | | Less Depressed (<26.1) | 31 | 6.581 (0.502) | |
| | GRU | More Depressed (>=26.1) | 24 | 4.375 (1.689) | $F=13.16$ ***p=.038*** |
| | | Less Depressed (<26.1) | 31 | 3.484 (1.288) | |

We could observe that CNN and GRU can distinguish the difference between the depressed person (M=6.000, SD=1.1216 / M=4.375, SD=1.689) and less depressed person (M=5.065, SD=2.065 / M=4.375, SD=3.484), respectively. The result can be interpreted as that GRU can evaluate whether the person is depressed or not. However, the result of CNN is unreasonable. It is well known that depressed people usually do not reveal their feelings; the evaluation score should be near 4.5 or are more negative than non-depressed; the score should be less than non-depressed.

## Conclusion

In this study, starting with a question, what we really want to find by sentiment analysis, we suggest a new framework to evaluate sentiment analysis model. By simple experiment with existing deep neural networks, which are CNN, LSTM, and GRU, we showed that although CNN performed the best in accuracy among other deep neural network algorithms (LSTM, GRU), when it is compared with human's sentiment, its outputs are not reasonable. The result implies that the accuracy of the model cannot reflect the psychological state of the person. Rather, GRU shows more explainable results related to the psychological states.

The contribution of this paper can be summarized as follows: Firstly, enlightening the relationship between computational models and psychological measurements. Recent sentiment analysis models should be able to explain not only whether the data is negative or positive but also whether the person is negative or positive. Second, this framework can be suggested as objective methods to evaluate emotional arousal of a person by what he/she said without using psychological measurements. If the model is trained well, it can be used to diagnose an emotional disorder. One of example could be in a school. The model could detect potential disorder by analyzing students' essays. This method does not need to use psychological measurement, which could be regarded that they are tested whether abnormal. The teenagers do not need to visit the mental clinic as well, which could be reluctant to them. The computational models are easy to be specialized on a specific domain, so we can build various models to predict not only emotionally expressive (the sum of PANAS), depressed (CES-D) but also other emotional states which are known to affect memory such as anxiety (Hertel & Brozovich, 2010).

In sentiment analysis, how we define the meaning of scores is an important crossroad to interpret the results. Specifically, 1 point can be interpreted in two ways: very negative or very impassive. Therefore, both valence and intensity have to be taken into consideration when drawing a conclusion with output scores. When scores were considered as valence, more expressive person has shown a tendency of thinking negatively, when compared to the less expressive person. However, unlike our background, a more depressed person has a tendency of saying positively. The unreasonable result can be explained when the scores were considered as intensity. More expressive people and less depressed people has low intensity when compared to less expressive people and more depressed people, respectively. The results correspond to our previous research which states that depressed people tend to overgeneralize their memory to avoid negative experience through affect regulation (Jo et al., 2016) as well as use the limited word and expressions than non-depressed (Jo et al., 2017). Moreover, the reason why CES-D is more correspond to sentiment analysis models than PANAS might be the characteristic of the measurements. Even though both PANAS and CES-D can influence memory retrieval, CES-D is more periodical measurement than PANAS. The depression gradually could have an influence on memory for a long time whereas affectivity is determined in a moment. Therefore, their writing style and memory might be smeared with depression (Thomas & Duke, 2007; Rude et al., 2004)

We have wondered that scoring from 1 to 9 is too detailed to distinguish between levels of emotion and would result in a poor accuracy. For instance, when taking two people – one who have scored 1 point and the other who have scored 2 points – into consideration, the difference between two test subjects should be observed but not much different in real.

Limitation and tentative work plan are still as follows: the number of training data might be insufficient for our model to classify sentiments in better performance. Data imbalance can be a reason. Secondly, the movie review data is not reliable because some companies employ part-time jobs to fabricate review scores of their movies. Although we ignored the 10 point reviews to reduce the data impurity as well as data imbalance, it does not mean the data is cleaned. Plus, transfer learning might be an inappropriate approach in this case - movie review to text summary. Considering our model accuracy, we should find other complex sentiment analysis architecture as a future work.